\documentclass[a4paper,12pt,oneside,reqno]{amsart}
\usepackage{graphicx}
\usepackage[hidelinks,colorlinks=true,linkcolor=blue,citecolor=blue]{hyperref}
\usepackage{caption}
\usepackage{algorithmic}
\usepackage{algorithm}
\usepackage{cite}
\usepackage{changepage}
\usepackage{pgfplots}
\usepackage{pgfplotstable}
\usepackage{booktabs}
\usepackage{array}
\usepackage{colortbl}
\usepackage{microtype}
\usepackage{adjustbox}
\usepackage{caption}
\usepackage{lmodern}
\usepackage{tikz}
\usepackage{subcaption}
\usepackage{cleveref}
\usepackage{siunitx}
\usepackage{amsthm}
\usepackage{xcolor}

\makeatletter
\def\subsection{\@startsection{subsection}{2}%
  \z@{.5\linespacing\@plus.7\linespacing}
{.5\baselineskip}%
  {\normalfont\scshape}%
}
\makeatother

\makeatletter
\def\subsubsection{\@startsection{subsubsection}{2}%
  \z@{.5\linespacing\@plus.7\linespacing}
{.5\baselineskip}%
  {\normalfont\scshape}%
}
\makeatother

\pgfplotsset{compat=newest}
\pgfplotsset{every axis legend/.append style={%
cells={anchor=west}}
}
\usetikzlibrary{arrows}
\tikzset{>=stealth'}

\pgfplotstableset{
  every head row/.style={before row=\toprule,after row=\midrule},
  every last row/.style={after row=\bottomrule},
  fixed,precision=2,
}

\begin{document}

\title{Hybrid Cuckoo Search Algorithm for the Minimum Dominating Set Problem}
\author[BELKACEM ZOUILEKH and Sadek BOUROUBI]{ \textbf{BELKACEM ZOUILEKH$^{1}$, Sadek BOUROUBI$^{2,*}$}\\
\MakeLowercase{\texttt{bzouilekh@usthb.dz}$^1$, \texttt{sbouroubi@usthb.dz}$^2$}  \vspace{3mm}\\\emph{$^{1,2\,}$LIFORCE L\MakeLowercase{aboratory}, \vspace{1mm}\\
USTHB, F\MakeLowercase{aculty of }M\MakeLowercase{athematics}, D\MakeLowercase{epartment of} O\MakeLowercase{perations }R\MakeLowercase{esearch},\vspace{1mm} \\ P.B. 32 E\MakeLowercase{l-Alia}, 16111, B\MakeLowercase{ab }E\MakeLowercase{zzouar}, A\MakeLowercase{lgiers}, A\MakeLowercase{lgeria.}}
}
\thanks{$^{*}$Corresponding author.}

\begin{abstract}
The notions of dominating sets of graphs began almost 400 years ago with the game of chess, which sparked the analysis of dominating sets of graphs, at first relatively loosely until the beginnings of the 1960s, when the issue was given mathematical description. It's among the most important problems in graph theory, as well as an NP-Complete problem that can't be solved in polynomial time. As a result, we describe a new hybrid cuckoo search technique to tackle the MDS problem in this work. Cuckoo search is a well-known metaheuristic famed for its capacity for exploring a large area of the search space, making it useful for diversification. However, to enhance performance, we incorporated intensification techniques in addition to the genetic crossover operator in the suggested approach.
The comparison of our method with the corresponding state-of-the-art techniques from the literature is presented in an exhaustive experimental test. The suggested algorithm outperforms the present state of the art, according to the obtained results.\\

\vspace{0.2cm}
\end{abstract}

\maketitle

\section{Introduction}\label{sec:introduction}
\begin{figure}[H]\centering
 \begin{adjustwidth}{-0.1\textwidth}{-0.2\textwidth}
  \includegraphics[width=.9\linewidth,height=\textheight,keepaspectratio]{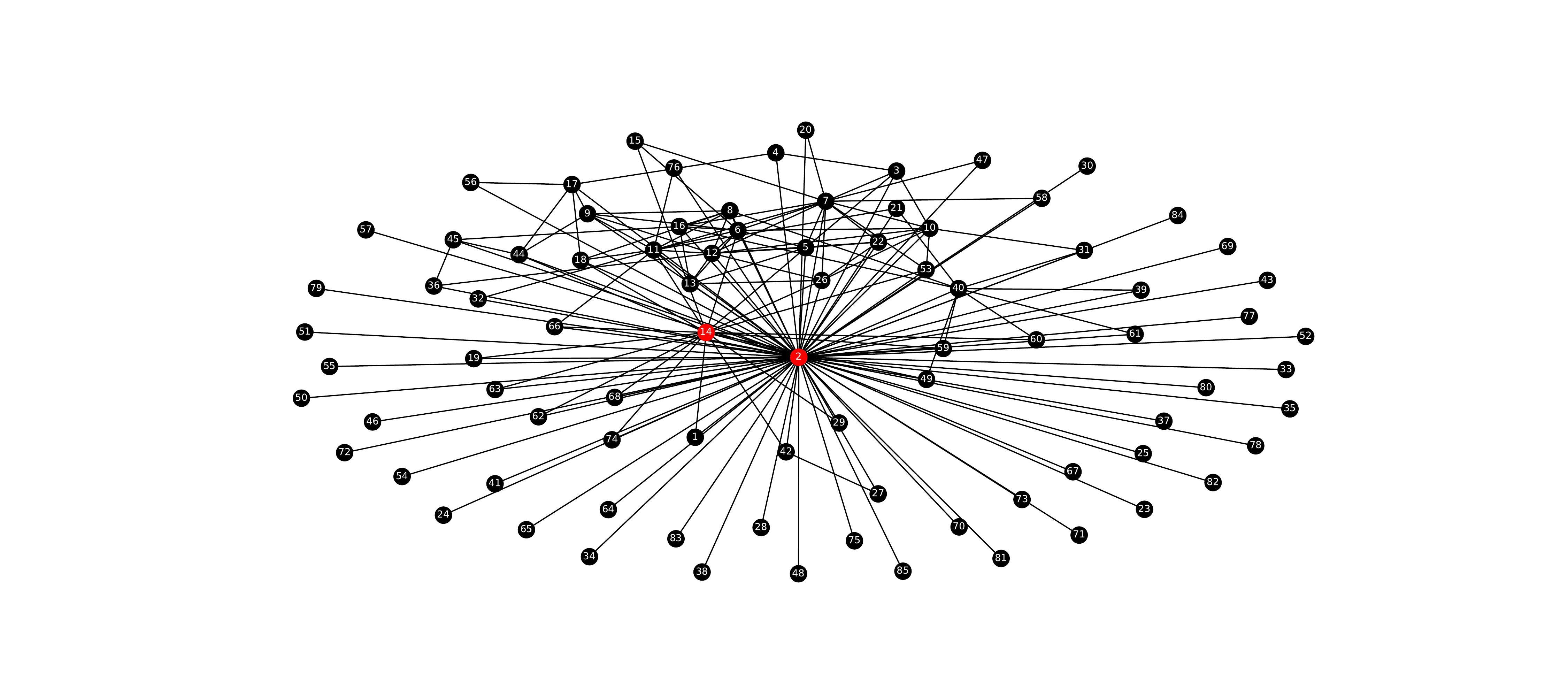}\end{adjustwidth}
  \caption{Representations of a real induced sub-graph from Twitter.}
  \label{fig:covid}
\end{figure}

Given an undirected graph $G = (V, E)$, a subset $S$  of $V$ is named a dominating set of $G$ if each vertex $v \in V$ is either an element of $S$ or is adjacent to an element of $S$. Such subset is called a dominating set of $G$ and we say $S$ dominates $G$, or $G$ is dominated by $S$. The minimum cardinality of a dominating set in $G$ is denoted by $\gamma(G)$. If $S$ dominates $G$ such as $|S|=\gamma(G)$ then $S$ is called a Minimum Dominating Set, a MDS for short \cite{haynes2013fundamentals}. As an example, Figure \ref{fig:covid} shows a subgraph of Twitter tweets with 85 vertices that spread specific 5G false news that is linked directly to the COVID-19 pandemic \cite{schroeder2021wico}, with nodes denoting Twitter users and edges representing follower relationships, The MDS obtained by our approach is highlighted in red. The original status publisher is represented by the vertices of the dominating set, which are marked in red. If we assume the simplest scenario, followers, marked in blue, can not retweet and content can not spread out of followers, only 2 users could probably spread misinformation among the other 83 users.

The minimum dominating sets and their variations, such as minimum connected dominating set and minimum weighted dominating set, have pervasive applications in a wide range of disciplines, besides Routing in ad hoc wireless networks \cite{dai2004extended,wu2001dominating}, Sensor Networks, and MANETs \cite{blum2004connected}. In addition, the MDS can be employed to determine the main subject sentences for document summary via sentence network design \cite{shen2010multi,xu2016generalized}. Furthermore, the MDS may be used to model and investigate the positive impact in social networks \cite{dinh2014approximability,daliri2018minimum,wang2011positive,wang2014finding}. Moreover, controlling the spread of epidemics \cite{takaguchi2014suppressing} also early diagnosis and control of epidemic spreadings in various areas of human society, such as Virus spread in computer systems, misinformation (as shown Figure \ref{fig:covid}), and content diffusion via social media \cite{zhao2020minimum}. MDS could be found in biological aspects such as interacting protein networks \cite{wuchty2014controllability} and biological network analysis \cite{nacher2016minimum}.\\

In this paper, we address the minimum dominating set problem by proposing a new Hybrid Cuckoo Search Algorithm, henceforth called \textbf{HCSA-MDS}, that combines the cuckoo search algorithm and the genetic crossover operator with intensification schemes as well as repairing and cleaning, to achieve effective exploration and exploitation.
\subsection{Paper structure}
The remainder of this paper is partitioned into many sections. In the \Cref{sec:Related works} that follows, we present some linked works, including exact algorithms and heuristics, applied to solving the MDS problem. \Cref{sec:CSA} goes into detail about the Cuckoo Search Algorithm. The Hybrid CSA algorithm is then presented and described in \Cref{sec:HCSA}. \Cref{sec:Experimental}, summarizes the results of the computation. The paper is finally concluded in the final \Cref{sec:Conclusion}.
\section{Related works}\label{sec:Related works}
The MDS problem is an NP-Hard combinatorial problem \cite{garey1979computers}, it has been thoroughly investigated, using exact (exponential-time) techniques. Lately, several authors have independently developed exact algorithms that solve the MDS problem in a graph with $n$ vertices better than the trivial $\mathcal{O}(2^{n})$ time brute force methodology, which merely searches through all possible case subsets of $V$.\\

Initially, Fomin et al. proposed a nice exact algorithm based on the search space restriction that breaks the natural $2^{n}$ barrier for the MDS problem to an $\mathcal{O}(1.9379^{n})$ time algorithm on arbitrary graphs and lower time in some graph classes \cite{fomin2004exact}. Then came various efforts on decreasing the time complexity of dominating sets, such as Ingo Schiermeyer who design an $\mathcal{O}(1.8899^{n})$ running time algorithm which is based on matching techniques to limit the search space \cite{schiermeyer2008efficiency}. After that, Grandoni decreased the running time to $\mathcal{O}(1.81^{n})$ \cite{grandoni2006note}. Then there were the works of Fomin et al.\cite{fomin2005some}, van Rooij and Bodlaender \cite{van2008design} and van Rooij et al. \cite{van2009inclusion}, which lowered the running time to around $\mathcal{O}(1.52^{n})$ more or less and in \cite{van2011exact}, van Rooij and Bodlaender reduced the running time to $\mathcal{O}(1.4969^{n})$. Currently, according to our information, the best exact algorithm for the MDS problem performs in $\mathcal{O}(1.4864^{n})$ time and polynomial space, and it is constructed through the measure and conquer approach \cite{iwata2011faster} by Y. Iwata.\\

Unfortunately, the exact techniques that execute at an exponential scale are only possible for limited-size networks, severely limiting their effective uses. As a result, scientific researchers have mainly concentrated on stochastic computational heuristics and lately metaheuristics. Hence, many heuristic approaches have been adopted in the state of the art to handle the MDS problem, such as \cite{parekh1991analysis,wan2003simple,alkhalifah2004genetic,misra2009minimum}. Moreover, L. A. Sanchis \cite{sanchis2002experimental} performed experimental research on different heuristic approaches in this perspective. He had thoroughly investigated many greedy methods for the MDS problem. After that, Ho et al.\cite{ho2006enhanced} presented ACO-TS, an improved Ant Colony Optimization metaheuristic that integrates a technique for stimulating the building of different solutions based on a concept adopted from genetic algorithms called tournament selection. Subsequently, in \cite{hedar2010hybrid}, Hedar and Ismail presented a hybrid genetic algorithm referred to as HGA-MDS that employs a local search which is characterized by three intensification techniques. Furthermore, in \cite{hedar2012simulated}, also Hedar and Ismail suggested a SAMDS metaheuristic addressing the MDS problem by employing a Stochastic Local Search (SLS) algorithm for strengthening a solution by looking for a stronger one in its near area. The SLS is enhanced by using a simulated annealing process. Recently, Abed and Rais \cite{abed2017hybrid}, introduced a hybrid population-based technique known as the Hybrid Bat Algorithm,  which is rooted in microbat bio-sonar characteristics and simulated annealing. The fact that SA is efficient in exploitation and the bat algorithm has a high capacity for the exploration of large regions in the search space helps to ensure a good balance between intensification and diversification in the search methodology.
\section{Cuckoo Search Algorithm}\label{sec:CSA}
The Cuckoo Search Algorithm (CSA), based on fascinating breeding behavior of particular cuckoo species, such as brood parasitism, is one of the most recently developed metaheuristics \cite{gandomi2013cuckoo}.\\

\begin{figure}[H]\centering
  \includegraphics[width=7cm, height=4.5cm,keepaspectratio]{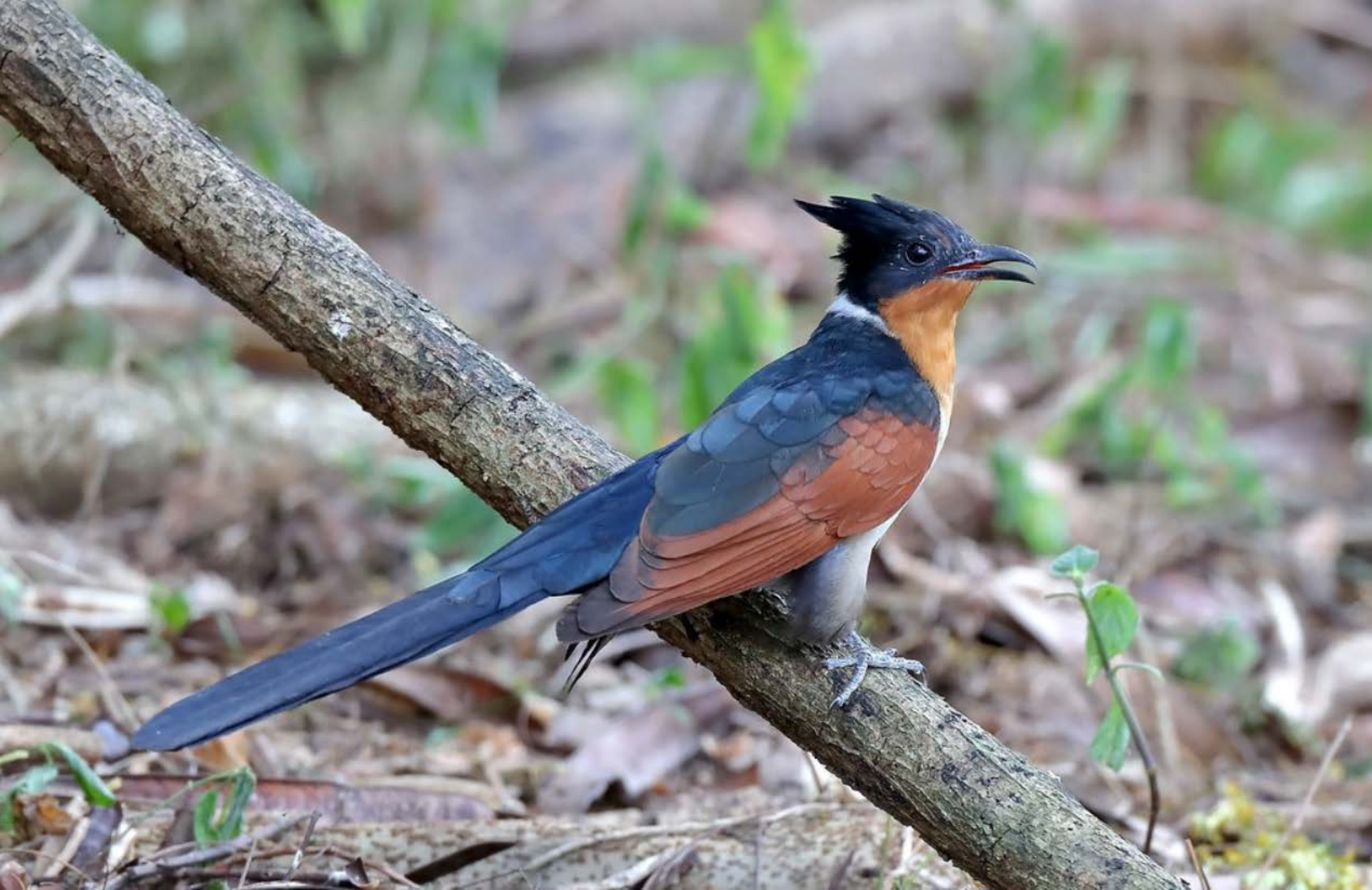}
  \caption{Chestnut-winged Cuckoo.}
  \label{fig:cuckoo}
\end{figure}

Cuckoos are intriguing birds (see Figure \ref{fig:cuckoo}), not just for their wonderful calls, but also for their aggressive breeding tactics. Some cuckoo species, such as the ani and guira, deposit their eggs in communal nests. However, they may remove other eggs to maximize the likelihood of their eggs hatching\cite{payne2005cuckoos,payne2020chestnut}.

To describe Cuckoo Search for clarity, Yang and Deb use the following three idealized principles\cite{yang2009cuckoo}:
\begin{enumerate}
    \item One egg is laid by each cuckoo at a time, and it deposits it in a nest that is selected at random.
    \item The best nests with the highest quality eggs (solutions with the highest fitness) will be passed on to future generations.
    \item The number of host nests is fixed, and a host has a probability of discovering an alien egg of $P_a \in [0, 1]$. In this scenario, the host bird can either discard the egg or leave the nest and create a new one at a different site.
\end{enumerate}

A solution's fitness can simply be proportional to the objective function's value. A cuckoo egg signifies a new solution, and each egg in a nest indicates a solution. The goal is to replace the poor solutions in the nests with new and maybe improved solutions (cuckoos).
This approach can be expanded to a more sophisticated scenario of several eggs representing a set of solutions in each nest. Therefore, we will take the simplest approach possible, with each nest containing only one egg.
The CSA pseudo-code is shown in Algorithm \ref{alg:CSA} based on these criteria.\\

\begin{algorithm}
\caption{Cuckoo Search Algorithm: CSA}
\hspace*{\algorithmicindent} \textbf{Input:} Problem instance  $S$ \\
\hspace*{\algorithmicindent} \textbf{Output:} the best possible solution  of $S$
\label{alg:CSA}
\begin{algorithmic}[1]
\STATE Initialize the population of $m$ host nests (solutions) $s = (s_1, \ldots, s_m)$;
\STATE $t = 0$;
\STATE $maxGen =$ Maximum number of generations;
\WHILE{$t\leq maxGen$}
   	\STATE Select a cuckoo (say $x_i$) at random using L\'{e}vy flights and evaluate its fitness $f(x_i)$;
	\STATE Choose one of $n$ (say $y_i$) nests at random;
	\IF{$f(x_i) > f(y_i)$}
		\STATE replace the old nest $x_i$ by the new one $y_i$
	\ENDIF
\STATE A portion of the worst nests $p_a$ are removed and replaced with new ones;
\STATE Sort the solutions and choose the best one;
\STATE Refresh the current best solution;
\STATE $t = t+1$;
\ENDWHILE
\end{algorithmic}
\end{algorithm}

After generating the initial population, CSA generates new solutions $x^{i+1}$ associated to each cuckoo $i$ in each iteration $t$ by a random walk via L\'{e}vy flight:\\
\begin{equation}
x_{i}^{t+1}=x_{i}^{t}+\alpha \oplus L\acute{e}vy(\lambda )
\end{equation}

Several researchers have revealed that the flight behavior of many animals and insects exhibits typical L\'{e}vy flights characteristics  \cite{brown2007levy,reynolds2007free,pavlyukevich2007levy,pavlyukevich2007cooling}.\\

L\'{e}vy flights are a type of random walk named after the French mathematician Paul L\'{e}vy \cite{brown2007levy}, in which $\alpha$  in the equation above denotes the step size which should correspond to the problem's interests (most of the time $\alpha = 1$) and the term product $\oplus$ denotes entrywise multiplications.
 The step lengths are selected from a probability distribution with a power law tail. L\'{e}vy distributions, or stable distributions, are the names given to these probability distributions.
\begin{equation}
L\acute{e}vy \sim u=l^{\lambda},
\end{equation}
where $1<\lambda\leq3$ and $l$ is the flight length \cite{viswanathan1999optimizing}.\\

The initial purpose of CSA and L\'{e}vy flights was to solve continuous optimization problems. Yang and Deb \cite{yang2009cuckoo} show the outperform and the robustness of CSA over GA and PSO, because of its larger search space exploration capacity and fewer parameters to fine-tune than other algorithms. There is just one parameter $P_a$, separate from the population size $N$, CSA has been widely used and shown promising efficiency in a variety of optimization and computational intelligence applications\cite{yang2014cuckoo}.\\

On the other hand, Tein and Ramli suggested a discrete cuckoo search algorithm to handle nurse scheduling problems \cite{lim2010recent}. While Boumedine and Bouroubi proposed a discrete hybrid cuckoo search algorithm to solve the protein folding problem \cite{boumedine2022protein}.\\

In this paper, we present a discrete hybrid CS-based method, called the Hybrid Cuckoo Search Algorithm for minimum dominating set (HCSA-MDS), to solve the MDS problem, which is one of the most difficult combinatorial optimization problems. The suggested discrete HCSA-MDS employs an adaptive L\'{e}vy flight for the MDS problem.
\section{Hybrid CSA for the MDS problem}\label{sec:HCSA}
We describe our approach to tackling the MDS problem in the following section. Our algorithm takes as input a problem instance $G = (V, E)$, in which $G$ is an undirected, connected graph, $V$ is a set of vertices, and $E$ is a set of edges.\\

\begin{center}
\begin{figure}
  \includegraphics[width=22cm,height=20cm,keepaspectratio]{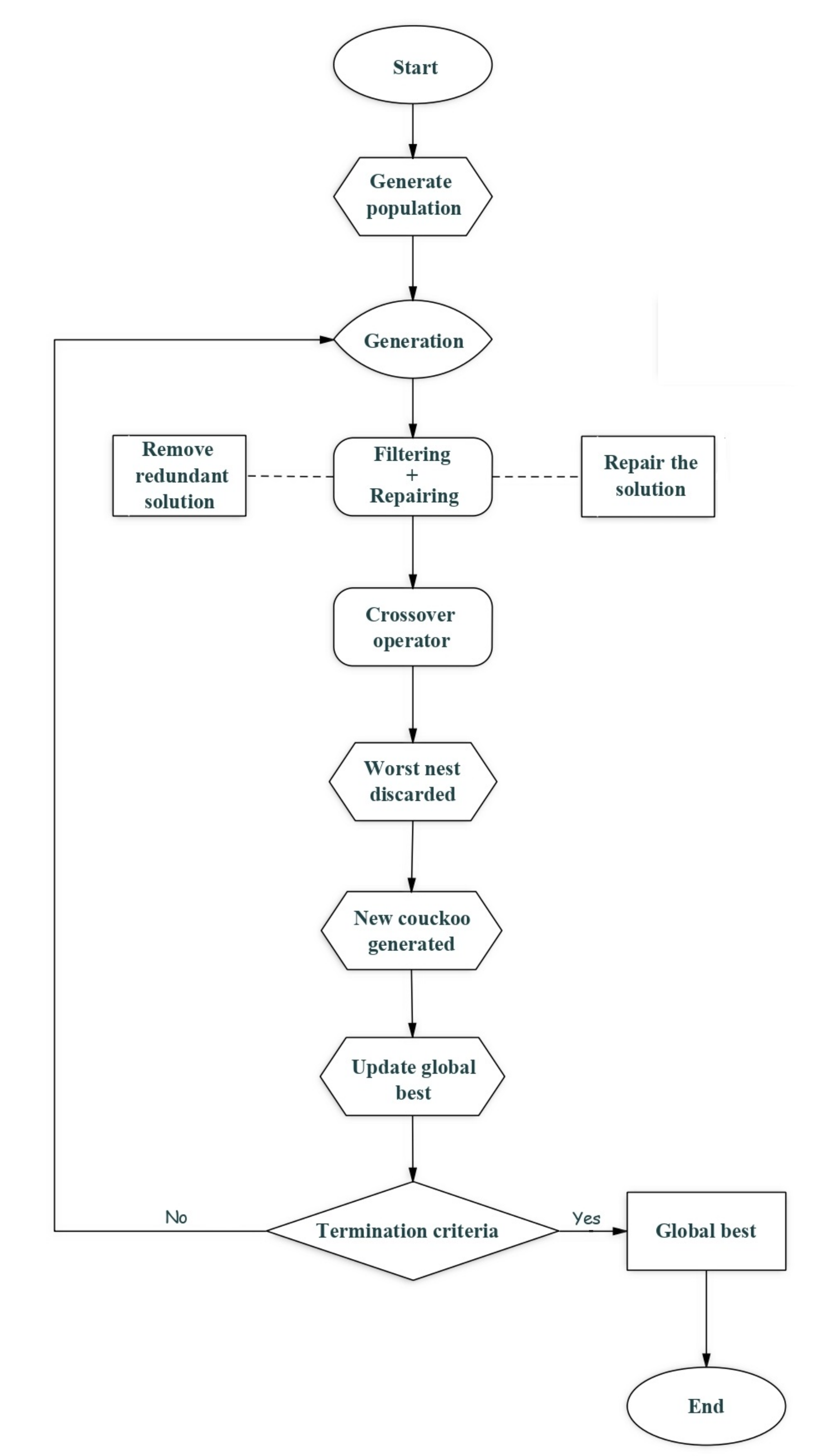}
  \caption{HCSA-MDS flowchart.}
  \label{fig:diagram}
\end{figure}
\end{center}

The HCSA-MDS combines the cuckoo search algorithm with the genetic crossover operator and the cleaning with repairing technique to exploit the solutions. While L\'{e}vy flights, with its ability to explore new regions in the search space, is a particularly useful strategy in the diversification phase. The goal of this hybridization is to establish a proper balance between search exploration and exploitation. The main structure is shown in Figure \ref{fig:diagram}.\\

We first discuss the HCSA-MDS components before fully stating the HCSA-MDS algorithm \ref{alg:HCSA-MDS}.
\subsection{Solution and population representation}
The HCSA-MDS algorithm starts with a population $P$ that contains a set of $m$ nests (solutions), some of which are non-feasible. As a result, a $0-1$ vector with a dimension equal to the number of nodes in the graph $G = (V, E)$ represents a solution $x$ in $P$. The ith node in $G$ is part of the dominating set if the vector's ith element has a value of 1. While the ith vector's item has a value of 0, this indicates that a node is not part of the dominant set. Table \ref{fig:binray-rep} illustrates the binary representation of the solution of the Example \ref{fig:covid} presented in \Cref{sec:introduction}.

\begin{table}[H]\centering
\begin{tabular}{ccccccccc}
\hline
1 & 2 & 3 & $\cdots$ & 13 & 14 & 15 & $\cdots$ & 85 \\ \hline
0 & \textcolor{red}{1} & 0 & $\cdots$ & 0 & \textcolor{red}{1} & 0 & $\cdots$ & 0 \\ \hline
\end{tabular}
\caption{Binary representation of the best found MDS}
\label{fig:binray-rep}
\end{table}

\subsection{A new cuckoo egg's production using the crossover operator}
The crossover operator is one of the genetic operators that applies to two parents (solutions) and then chooses at random any of the crossover points $p_h$ , $h\in \lbrace 0,1,...,n-1 \rbrace$ \cite{umbarkar2015crossover}. Two offspring (new solutions) are created by joining the parents at the crossover point for the next generation. However, for any solution from population $P$ that we consider the first parent, we randomly choose a second parent from the current population. Then, the one-point crossover operation is used to generate two offsprings. In the proposed algorithm, only the best offspring are chosen for the next stage, see the example in the following Table \ref{table:crossover}:
\small
\begin{table}[H]\centering
\begin{tabular}{|c|c|c|c|c|}
\hline
Parents & Binary representation &  & Offsprings & Binary representation \\
\cline{1-2}\cline{4-5}
First & \textcolor{blue}{1010}$\mid $\textcolor{red}{10110} & $\mathbf{\Rightarrow}$ & First & \textcolor{blue}{1010}$\mid $%
\textcolor{orange}{10010} \\ \cline{1-2}\cline{4-5}
Second & \textcolor{green}{1111}$\mid $\textcolor{orange}{10010} &  & Second & \textcolor{green}{1111}$\mid $\textcolor{red}{10110} \\ \hline
\end{tabular}
\caption{An illustrative example of a crossover operator with a single point-crossover.}
\label{table:crossover}
\end{table}
In the previous example, the two offsprings are generated by combining the first and the second parents at the fourth position choosed randomly. Furthermore, in the example, the crossover generates two new solutions for which, in terms of minimum dominating set, the first one outperforms its parents and even its sibling.
\subsection{Filtering procedure}
For intensification, HCSA-MDS employs a filtering mechanism that cleans the MDS solutions by removing part of the redundant nodes. Cleaning attempts to limit the dominating set represented by the solution while maintaining its coverage \cite{hedar2010hybrid}. The Filtering procedure is shown in \mbox{Algorithm \ref{alg:FP}:}
\begin{algorithm}[H]
\caption{Filtering procedure}
\hspace*{\algorithmicindent} \textbf{Input:} Solution $S$. \\
\hspace*{\algorithmicindent} \textbf{Output:} Filtered solution.
\label{alg:FP}
\begin{algorithmic}[1]
  	\FOR{each node $x$ in $S$}
  		\IF{$x = 1$}
  			\STATE Set $x = 0$;
  			\STATE compute the new fitness value of the solution $S$;
  			\IF{ the new fitness values is increased }
  				\STATE update the solution $S$ and set $x = 0$;
  			\ELSE
  				\STATE reset $x = 1$;
  			\ENDIF
  		\ENDIF
  	\ENDFOR
\end{algorithmic}
\end{algorithm}
\subsection{Reparation procedure}
In this phase, we fix the solution $S$ that does not cover all nodes in this procedure. First we check whether the solution is not a dominating set, then the next vertex to be added to the set of dominators is always picked from vertices that are not in $S$ with the maximum degree. The reparation process stops when $S$ is a dominating set, as shown in Algorithm \ref{alg:Reparation}.
\begin{algorithm}[H]
\caption{Reparation procedure}
\label{alg:Reparation}
\hspace*{\algorithmicindent} \textbf{Input:} Solution $S$. \\
\hspace*{\algorithmicindent} \textbf{Output:} Repaired solution.
\begin{algorithmic}[1]
   	\IF {there exist Non-covered nodes in the solution $S$}
   		\STATE Select a node $x$ with maximum degree in non-covered nodes set;
   		\STATE Set $x = 1$ and update the solution $S$;
   		\STATE Remove $x$ from non-covered nodes and its neighbours;
    \ENDIF
\end{algorithmic}
\end{algorithm}
\subsection{Construction of new solution via L\'{e}vy flights}
Fruit flies or Drosophila melanogaster, investigate their environment utilizing a sequence of straight flight pathways interrupted by a sharp $\ang{90}$ rotation, resulting in an L\'{e}vy-flying-style discontinuous free exploration pattern on a large scale, according to Reynolds and Frye \cite{reynolds2007free}. The basic features of L\'{e}vy flights can also be seen in studies of human behavior, such as foraging movement patterns that are used by human foragers \cite{brown2007levy}.\\

Therefore, because of the step length features, L\'{e}vy flights are more efficient in exploring the search space. We link the step length (i.e., the length of the subset) with the value created by the L\'{e}vy flights to employ this technique for MDS. For example, if $s$ is an $n$-dimensional solution, we divide the interval $\left[ 0,1\right]$ into $m$ subintervals and define the step length($step_{length}$) as stated below:
\begin{table}[H]\centering
\bgroup
\def\arraystretch{1.7}
\begin{center}
\begin{tabular}{|c|c|c|c|c|}
\hline
$\theta $ & $\left[ 0,\frac{1}{m}\right[ $ & $\left[ \frac{1}{m},\frac{2}{m}%
\right[ $ & ... & $\left[ \frac{m-1}{m},1\right] $ \\ \hline
$step_{length}$ & $\left[ 1,\frac{\max_{l}}{m}\right[ $ & $\left[ \frac{%
\max_{l}}{m},\frac{2\max_{l}}{m}\right[ $ & ... & $\left[ \frac{(m-1)\max_{l}%
}{m},\max_{l}\right] $ \\ \hline
\end{tabular}%
\end{center}
\egroup
\end{table}
\noindent where $\theta $ is the L\'{e}vy flights value acquired and $\max_{l} = \frac{n}{h}$, $h\in \left\{ 1,2,3,\ldots,n\right\}$ is the maximum subset we are able to invert. Let $i$ be the length of the subset chosen randomly from the interval generated by L\'{e}vy's flight value. Then, from the $n$-dimensional solution, we choose a random position between $1$ and $n-i$ from which the $i$-inversion begins.
\subsection{HCSA-MDS Algorithm}

\footnotesize
\begin{algorithm}[H]	
\caption{HCSA-MDS}
\label{alg:HCSA-MDS}
\hspace*{\algorithmicindent} \textbf{Input:} A graph $G = (V,E)$. \\
\hspace*{\algorithmicindent} \textbf{Output:} Best Dominating Set.
\begin{algorithmic}[1]
  \STATE Initialization: Generate an initial population $P$ of $m$ feasible solutions,\\
$S = (x_{i}, \ldots , x_{m})$;
  \STATE $Gen = 0$;
  \WHILE {$Gen \leq Max$ (maximum number of generations)}
  	\STATE $i = 1$;
  	\WHILE {$i \leq m$}
  		\STATE Filter repaired solution to improve its quality;
  		\STATE Select a random solution $x_{j}$ from $P$ as second parent;
		\STATE Produce a new cuckoo eggs $y_{j}$ using the crossover operator  to the selected parents $x_{i}$ and $x_{j}$;
		\IF{$f(y_{j}) \geq f(x_{i})$}
			\STATE Replace $x_{i}$ by the new produced solution $y_{j}$;	
		\ENDIF
  	\ENDWHILE
  	\STATE  The worst solutions are removed from $P$ and new ones are generated via L\'{e}vy flights  proportional to  $P_a \in \left[ 0,1 \right] $;
  	
  	\STATE Rank the solutions from the best to the worst and find the best one;
  	\STATE Update the global optimal solution;
  \ENDWHILE
\end{algorithmic}
\end{algorithm}
\section{Experimental results}\label{sec:Experimental}
HCSA-MDS effectiveness is evaluated against a collection of well-known efficient algorithms for the Minimum Dominating Set problem that has been reported in the literature. The rest of this section will be as follows. First, all benchmarks used are presented. Then, we compare the obtained results by the proposed algorithm with the state-of-the-art approaches.\\

To determine the relevance of the differences between the approaches results and HCSA-MDS Wilcoxon's signed-rank test is used. It's a nonparametric test method for analyzing matched-pair data. In addition, we employ Critical Difference (CD) diagrams that help to clarify things.\\

The Wilcoxon test is a powerful tool for detecting substantial variations between two algorithms' performance, which will assist us in:
\begin{enumerate}

\item Determine which individual of a pair is "greater than" i.e., determine the sign of any pair's distinction.
\item Compare and contrast the differences.

\end{enumerate}
The Wilcoxon signed-rank test is used for the evaluation of matched-pair statistics. The null hypothesis in the Wilcoxon's signed-rank test is whether the probability distributions of the first and second data are equivalent, denoted $\mathcal{H}_0$ which means, the sum of the good ranks (the value of the first algorithm is greater than the second algorithm) is equal to the sum of the bad ranks (the value of the second algorithm is bigger than the first algorithm). However, if the sum of the good ranks differs from the sum of the bad ones, we reject $\mathcal{H}_0$ in favor of the alternative hypothesis, symbolized by $\mathcal{H}_1$, that the two algorithms differ \cite{sheskin2003handbook}.\\

When two scores achieved by any pair of algorithms are equal, which is signified by ties, all tied cases are removed from the analysis, reducing $N$, the number of matched pairs. This hypothesis can be evaluated using statistics based on intrapair differences and a null hypothesis refusal area equivalent to or less than $\alpha = 0.05$ for a two-tailed test where the hypothesis is that the frequency for which the two signs appear will differ greatly.\\

Furthermore, Wilcoxon's statistic uses these ranks to build a $Z$ score, which will be compared to a typical normal distribution to determine whether there are any variations between the algorithms. The $p$ value is then evaluated by comparing it to the $0.05$ significance level. We consider the null hypothesis if it is greater or equal to 0.05, for which the compared two algorithms do not differ. In contrast, if it is less than 0.05, we consider the alternative hypothesis, which is that the two algorithms differ and then one outperforms the other \cite{toothaker1989book}.\\

Critical difference (CD) diagrams, initially described in \cite{demsar2006statistical}, were used to rank algorithms. They are a helpful tool for evaluating various algorithms.
Before plotting any diagram, first use the non-parametric Friedman test, which analyzes each data set's algorithms independently to determine the average ranks of algorithms to see whether there are any substantial differences.
When the null hypothesis that all perform equally is well rejected, we conduct a post-hoc analysis. A Wilcoxon signed-rank test is used in the second phase to determine whether each pair of algorithms differs significantly \cite{benavoli2016should}. We must use Holm's approach to adapt Wilcoxon's test because we are evaluating several hypotheses. Each considered algorithm is ranked on the horizontal axis in these diagram plots. The algorithm with the best performance receives a rank close to 1, the second receives a bigger rank, and so on. Strong horizontal bars connect the corresponding algorithms, for which we cannot differentiate from the Holm-adjusted Wilcoxon's test, regarded as statistically equal.
\subsection{Data Sets}\label{sec:data sets}
The following two benchmarks are widely used in the literature for the MDS problem:
\begin{enumerate}
\item 42 random geometric graphs with upwards to 400 nodes, previously given in \cite{hedar2012simulated}. The authors used the generating instructions from \cite{bettstetter2002minimum,ho2006enhanced} to create the networks. Each network is created by inserting $ n $ nodes, see column \textbf{N\textsuperscript{o} of nodes} in a $M \times M$ area at random according to the uniform distribution as shown in the column \textbf{Area} in Table \ref{table:Table1}. Several graph instances were generated for each network based on the range parameter listed in column \textbf{Range} in Table \ref{table:Table1}. They followed the instructions in \cite{bettstetter2002minimum} to guarantee that the graph formed a connected graph. Moreover, the number of graph instances per network was also included in the column \textbf{N\textsuperscript{o} of instances}.

\begin{table}[H]\centering
\begin{adjustbox}{width=0.8\linewidth,center}
\begin{tabular}{ccccc}
\hline
Network Id & N\textsuperscript{o} of nodes ($n$) & Range & Area (A) & N\textsuperscript{o} of instances \\ \hline
N1$_{A}^{n}$ & 80 & 60-120 & 400$\times $400 & 7 \\
N2$_{A}^{n}$ & 100 & 80-120 & 600$\times $600 & 5 \\
N3$_{A}^{n}$ & 200 & 70-120 & 700$\times $700 & 6 \\
N4$_{A}^{n}$ & 200 & 100-160 & 1000$\times $1000 & 7 \\
N5$_{A}^{n}$ & 250 & 130-160 & 1500$\times $1500 & 4 \\
N6$_{A}^{n}$ & 300 & 180-220 & 2000$\times $2000 & 5 \\
N7$_{A}^{n}$ & 350 & 200-230 & 2500$\times $2500 & 4 \\
N8$_{A}^{n}$ & 400 & 210-240 & 3000$\times $3000 & 4 \\ \hline
\end{tabular}
\end{adjustbox}
\caption{First benshmark.}
\label{table:Table1}
\end{table}

\item  21 graphs of 400 and 800 vertices, provided from \cite{hedar2012simulated}. In this benchmark set, the optimal dominating set is known. The authors employed the approach described in \cite{sanchis2002experimental} to generate graphs with known domination numbers and provided densities henceforth denoted $\mathbf{N_{d,p}^{n}}$, where $n$ is the number of nodes, $d$ the domination number, and $p$ is the probability for edge creation. The detailed procedure is initially presented in \cite{sanchis2002experimental}.\\

In this benchmark, we used two graphs of 400 and 800 vertices with different densities (0.1,0.3 and 0.5) and different domination numbers as presented in Table \ref{table:Table2}.
\begin{table}[H]\centering
\begin{adjustbox}{width=0.8\linewidth,center}
\begin{tabular}{ccccc}
\hline
Network & $n$ & $p$ & $d$ & N\textsuperscript{o} of graph instance \\ \hline
$N_{d,0.1}^{400}$ & $400$ & $0.1$ & $8,11,14,18,23$ & $5$ \\ \hline
$N_{d,0.3}^{400}$ & $400$ & $0.3$ & $3,5,8,11,14$ & $5$ \\ \hline
$N_{d,0.5}^{400}$ & $400$ & $0.5$ & $3,5,8,11$ & $4$ \\ \hline
$N_{d,0.1}^{800}$ & $800$ & $0.1$ & $11,14,22$ & $3$ \\ \hline
$N_{d,0.3}^{800}$ & $800$ & $0.3$ & $3,5$ & $2$ \\ \hline
$N_{d,0.5}^{800}$ & $800$ & $0.5$ & $3,6$ & $2$ \\ \hline
\end{tabular}
\end{adjustbox}
\caption{Second benchmark.}
\label{table:Table2}
\end{table}
\end{enumerate}

Note that there are 63 graphs in both benchmark sets.
\subsection{Tuning algorithm parameters}\label{sec:tuning parameters}
To conduct our experiment, we set the cuckoo search parameters respecting the common parameters used in the literature to address various optimization issues \cite{yang2009cuckoo}. Table \ref{table:Table3} shows the HCSA-MDS parameter values that produce the best performance. The first parameter $N$ is the population size, and the next parameter is maxGen, which represents the number of generations. However, the probability of the host bird discovering a cuckoo egg is $P_{a}$. Finally, the cuckoo's step length parameters are $\gamma$ and $\alpha$.
\begin{table}[H]\centering
\begin{tabular}{cc}
\hline
Parameter & Value \\ \hline
$N$ & $50$ \\
$MaxGen$ & $500$ \\
$P_{a}$ & $0.25$ \\
$\gamma $ & $3/2$ \\
$\alpha $ & $1$ \\ \hline
\end{tabular}
\caption{HCSA-MDS Parameters.}
\label{table:Table3}
\end{table}

\subsection{The numerical performance}
Each benchmark was performed 10 times on every graph from the two literature data sets. However, for the second test case, we obtain the optimal dominant set from the first generation in all ten runs, except for four graphs that require more than 20 generations, so we generalize the number of generations up to 500 as stated in \Cref{sec:tuning parameters}. Python was used to implement our HCSA-MDS algorithm. Furthermore, all testing was carried out on a system with 16 Gigabytes of RAM and a 2.6 GHz Intel Core i5 CPU. The suggested HCSA-MDS performance was evaluated using the two groups of data sets presented in \Cref{sec:data sets}. Table \ref{table:firstbenchmark} shows the results of HCSA-MDS on 42 graph instances generated by the initial test cases. Tables \ref{table:N400} and \ref{table:N800} show the other HCSA-MDS performance on 21 graph instances generated by the second test scenarios.
\subsubsection{Comparative analysis of HCSA-MDS against the state-of-art approaches on the first benchmark}
We test the suggested HCSA-MDS on the first benchmark against HGA-MDS \cite{hedar2010hybrid}, SAMDS \cite{hedar2012simulated}, and HBA \cite{abed2017hybrid} for the purpose of proving the performance of the HCSA-MDS approach check Table \ref{table:firstbenchmark}.\\

\begin{table}
\begin{adjustbox}{width=1.5\linewidth,center}
\begin{tabular}{lllllllllllllll >{\bfseries}l lll}
\hline
&  &  & \multicolumn{3}{c}{HGA-MDS} &  & \multicolumn{3}{c}{SAMDS} &  &
\multicolumn{3}{c}{HBA} &  & \multicolumn{4}{c}{HCSA-MDS} \\
\cline{4-6}\cline{8-10}\cline{12-14}\cline{16-19}
Graph Id & Range &  & Best & Avg. & Std. &  & Best & Avg. & Std. &  & Best &
Avg. & Std. &  & Best & Avg. & Std. & Worst \\
\cline{1-2}\cline{4-6}\cline{8-10}\cline{12-14}\cline{16-19}
N1$_{400}^{80}$ & 60 &  & 15 & 15.95 & 0.39 &  & 15 & 16.35 & 0.67 &  & 16 &
16.00 & 0 &  & 14 & 14.00 & 0 & \textbf{14}$^{\star }$ \\
& 70 &  & 13 & 14.00 & 0.73 &  & 12 & 14.40 & 1.14 &  & 13 & 13.30 & 0.48 &
& 11 & 11.00 & 0 & \textbf{11}$^{\star }$	 \\
& 80 &  & 10 & 10.85 & 0.59 &  & 10 & 12.00 & 1.03 &  & 10 & 10.60 & 0.52 &
& 9 & 9.10 & 0.32 & 10$^{\star }$ \\
& 90 &  & \textbf{8} & 8.40 & 0.50 &  & \textbf{8} & 9.60 & 1.27 &  & 9 & 9.60 & 0.52 &  & 8 & 8.00 & 0 & \textbf{8}$^{\star }$ \\
& 100 &  & \textbf{7} & 8,20 & 0.52 &  & \textbf{7} & 9.10 & 0.97 &  & 8 & 8.40 & 0.52 &  & 7 &  7.00 & 0 & \textbf{7}$^{\star }$ \\
& 110 &  & \textbf{6} & 6.05 & 0.22 &  & \textbf{6} & 7.55 & 0.69 &  & \textbf{6} & 6.90 & 0.32 &  & 6 & 6.00 & 0 & \textbf{6}$^{\star }$ \\
& 120 &  & \textbf{5} & 5.95 & 0.39 &  & \textbf{5} & 7.00 & 1.08 &  & 6 & 6.70 & 0.48 &  & 5 & 5.20 & 0.42 & 6 \\ \hline
N2$_{600}^{100}$ & 80 &  & 19 & 19.55 & 0.85 &  & 18 & 20.55 & 1.23 &  & 19
& 19.00 & 0 &  & 15 & 15.90 & 0.32 & 16$^{\star }$ \\
& 90 &  & 16 & 17.20 & 0.95 &  & \textbf{15} & 17.65 & 1.73 &  & 18 & 18.00 & 0 &  &
15 & 15.00 & 0 & \textbf{15}$^{\star }$ \\
& 100 &  & 14 & 14.85 & 0.49 &  & \textbf{13} & 14.95 & 1.05 &  & 14 & 14.40 & 0.52 &
& 13 & 13.20 & 0.42 & 14$^{\star }$ \\
& 110 &  & \textbf{11} & 12.15 & 0.67 &  & \textbf{11} & 12.85 & 1.23 &  & 12 & 12.30 & 0.48 & & 11 & 11.00 & 0 & \textbf{11}$^{\star }$ \\
& 120 &  & 10 & 10.15 & 0.37 &  & \textbf{9} & 12.00 & 1.49 &  & 11 & 11.00 & 0 &  & 9 & 9.00 & 0 & \textbf{9}$^{\star }$ \\ \hline
N3$_{700}^{200}$ & 70 &  & 37 & 45.65 & 10.83 &  & 35 & 39.95 & 2.09 &  & 34
& 35.40 & 0.84 &  & 29 & 29.50 & 0.71 & 31$^{\star }$ \\
& 80 &  & 31 & 33.05 & 1.32 &  & 29 & 33.50 & 1.73 &  & 28 & 28.20 & 0.42 &
& 25 & 25.50 & 0.97 & 28$^{\star }$ \\
& 90 &  & 26 & 28.75 & 2.67 &  & 25 & 29,25 & 1.94 &  & 25 & 26.00 & 0.67 &
& 20 & 21.20 & 0.63 & 22$^{\star }$ \\
& 100 &  & 21 & 24.70 & 4.94 &  & 20 & 24.20 & 1.70 &  & 21 & 21.80 & 0.42 &
& 18 & 19.00 & 0.47 & 20$^{\star }$ \\
& 110 &  & 19 & 19.95 & 0.51 &  & 18 & 21.40 & 1.93 &  & 17 & 17.70 & 0.48 &
& 16 & 16.20 & 0.42 & 17$^{\star }$ \\
& 120 &  & 17 & 17.30 & 0.47 &  & 15 & 19.55 & 1.43 &  & 15 & 16.20 & 0.79 &
& 14 & 14.00 & 0 & \textbf{14}$^{\star }$ \\ \hline
N4$_{1000}^{200}$ & 100 &  & 39 & 45.25 & 5.57 &  & 38 & 41.35 & 1.87 &  & 36
& 36.70 & 0.48 &  & 29 & 29.70 & 0.82 & 31$^{\star }$ \\
& 110 &  & 35 & 37.35 & 2.06 &  & 33 & 37.20 & 2.44 &  & 30 & 30.80 & 0.42 &
& 26 & 26.40 & 0.52 & 27$^{\star }$ \\
& 120 &  & 27 & 28.90 & 1.77 &  & 26 & 30.10 & 1.45 &  & 26 & 26.30 & 0.48 &
& 23 & 23.10 & 0.32 & 24$^{\star }$ \\
& 130 &  & 26 & 27.30 & 1.13 &  & 25 & 28.15 & 1.42 &  & \textbf{23} & 24.20 & 0.63 &
& 23 & 23.10 & 0.32 & 24 \\
& 140 &  & 23 & 24.35 & 1.35 &  & 22 & 25.70 & 1.84 &  & 22 & 23.00 & 0.67 &
& 20 & 20.90 & 0.32 & 21$^{\star }$ \\
& 150 &  & 21 & 21.45 & 0.76 &  & 20 & 23.55 & 1.43 &  & 19 & 20.30 & 0.82 &
& 18 & 18.10 & 0.32 & 19$^{\star }$ \\
& 160 &  & 20 & 21.60 & 0.94 &  & 19 & 21.30 & 1.30 &  & 18 & 18.00 & 0 &  &
17 & 17.00 & 0 & \textbf{17}$^{\star }$ \\ \hline
N5$_{1500}^{250}$ & 130 &  & 55 & 75.45 & 24.01 &  & 51 & 56.05 & 2.68 &  &
49 & 50.00 & 0.67 &  & 37 & 38.10 & 0.99 & 40$^{\star }$ \\
& 140 &  & 48 & 59.75 & 12.78 &  & 46 & 48.65 & 1.35 &  & 42 & 42.60 & 0.70
&  & 40 & 41.10 & 0.57 & 42$^{\star }$ \\
& 150 &  & 44 & 48.30 & 3.34 &  & 41 & 44.75 & 1.71 &  & \textbf{37} & 37.90. & 0.32
&  & 37 & 37.60 & 0.52 & 38 \\
& 160 &  & 38 & 41.65 & 3.42 &  & 37 & 40.90 & 1.92 &  & 31 & 32.60 & 1.71 &
& 30 & 30.40 & 0.52 & 31$^{\star }$ \\ \hline
N6$_{2000}^{300}$ & 180 &  & 54 & 61.20 & 6.64 &  & 47 & 52.35 & 2.41 &  & 44
& 45.70 & 0.95 &  & 43 & 43.10 & 0.32 & 44$^{\star }$ \\
& 190 &  & 48 & 55.55 & 6.35 &  & 46 & 50.20 & 2.24 &  & 44 & 44.40 & 0.52 &
& 40 & 41.10 & 0.74 & 42$^{\star }$ \\
& 200 &  & 41 & 47.90 & 5.07 &  & 40 & 45.25 & 2.55 &  & 41 & 41.10 & 0.32 &
& 36 & 36.40 & 0.52 & 37$^{\star }$ \\
& 210 &  & 40 & 48.60 & 7.99 &  & 39 & 43.60 & 1.70 &  & 37 & 37.70 & 0.82 &
& 34 & 34.40 & 0.52 & 35$^{\star }$ \\
& 220 &  & 36 & 39.90 & 4.34 &  & 36 & 40.65 & 1.90 &  & 33 & 34.10 & 0.57 &
& 31 & 31.80 & 0.79 & 33$^{\star }$ \\ \hline
N7$_{2500}^{350}$ & 200 &  & 67 & 93.45 & 23.58 &  & 61 & 66.35 & 2.13 &  &
58 & 58.90 & 0.32 &  & 47 & 48.10 & 0.57 & 49$^{\star }$ \\
& 210 &  & 63 & 91.20 & 26.70 &  & 58 & 61.85 & 2.18 &  & 53 & 55.00 & 1.05 &
& 44 & 45.60 & 0.84 & 47$^{\star }$ \\
& 220 &  & 55 & 76.85 & 30.55 &  & 49 & 55.05 & 2.31 &  & 51 & 51.80 & 0.63
&  & 44 & 44.60 & 0.52 & 45$^{\star }$ \\
& 230 &  & 51 & 67.00 & 21.02 &  & 48 & 54.05 & 2.42 &  & 45 & 47.10 & 1.37
&  & 39 & 40.00 & 0.82 & 41$^{\star }$ \\ \hline
N8$_{3000}^{400}$ & 210 &  & 79 & 115.55 & 41.21 &  & 75 & 80.15 & 3.10 &  &
72 & 73.30 & 0.95 &  & 58 & 59.90 & 1.45 & 62$^{\star }$ \\
& 220 &  & 77 & 110.45 & 39.76 &  & 73 & 79.25 & 3.18 &  & 70 & 71.00 & 0.82 &
& 56 & 56.40 & 0.52 & 57$^{\star }$ \\
& 230 &  & 73 & 111.55 & 38.94 &  & 71 & 74.10 & 2.22 &  & 64 & 64.00 & 0 &  &
60 & 60.70 & 0.48 & 61$^{\star }$ \\
& 240 &  & 70 & 103.15 & 32.02 &  & 63 & 68.80 & 2.98 &  & 58 & 59.30 & 0.67
&  & 55 & 56.60 & 0.70 & 57$^{\star }$ \\ \hline
\end{tabular}
\end{adjustbox}
\caption{Performance comparison of various algorithms for the first benchmark sets.}
\label{table:firstbenchmark}
\end{table}

\noindent HCSA-MDS performance is given for each instance in terms of:
\begin{enumerate}
	\item The best solution is defined by the column denoted "\textbf{Best}", which defines the better solution identified by each approach, which is the minimum dominating set found in all runs for each graph instance.
	\item Average: This indicator is given by the column labeled "\textbf{Avg.}", and it represents the mean of the finest solution values discovered in the individual runs of each graph instance.
	\item The column labeled "\textbf{Std.}" represents standard deviation, which is a statistic that measures the dispersion of the best solutions identified in runs.
	\item The worst solution is defined by the column denoted "\textbf{Worst}", which is only for HCSA-MDS this measure represents the worst solution found in 10 runs for each graph instance.
	\item The number of times we have gotten to the optimal solution in the column labeled "\textbf{Opt. Rea.}" (see Table \ref{table:N400} and Table \ref{table:N800}), which represents the number of runs on each graph for which the solution (domination number) was reached.
\end{enumerate}

The following assessment can be drawn from the previous experiments (see Table \ref{table:firstbenchmark}):\\

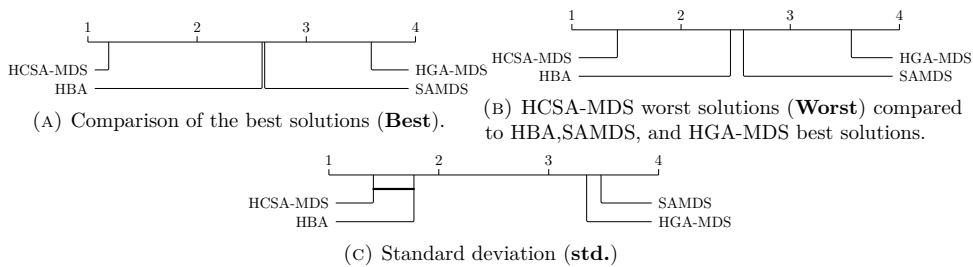
\begin{figure}[H]
\begin{adjustbox} {width=1\columnwidth,center}
\begin{subfigure}{.7\textwidth}
\begin{tikzpicture}[scale=0.72]
\begin{axis}[
  axis x line=center, axis y line=none, xmin=1, xmax=4, ymin=-3.5, ymax=0, scale only axis, height=4\baselineskip, width=\axisdefaultwidth, ticklabel style={anchor=south, yshift=1.33*\pgfkeysvalueof{/pgfplots/major tick length}, font=\small}, every tick/.style={yshift=.5*\pgfkeysvalueof{/pgfplots/major tick length}}, axis line style={-}, title style={yshift=\baselineskip}, xtick={1,2,3,4}, clip=false
]

\draw[semithick, rounded corners=1pt] (axis cs:1.1904761904761905, 0) |- (axis cs:1.0, -1.5) node[font=\small, fill=white, inner xsep=5pt, outer xsep=-5pt, anchor=east] {HCSA-MDS};

\draw[semithick, rounded corners=1pt] (axis cs:2.5952380952380953, 0) |- (axis cs:1.0, -2.5) node[font=\small, fill=white, inner xsep=5pt, outer xsep=-5pt, anchor=east] {HBA};

\draw[semithick, rounded corners=1pt] (axis cs:2.619047619047619, 0) |- (axis cs:4.0, -2.5) node[font=\small, fill=white, inner xsep=5pt, outer xsep=-5pt, anchor=west] {SAMDS};

\draw[semithick, rounded corners=1pt] (axis cs:3.5952380952380953, 0) |- (axis cs:4.0, -1.5) node[font=\small, fill=white, inner xsep=5pt, outer xsep=-5pt, anchor=west] {HGA-MDS};

\end{axis}
\end{tikzpicture}
\caption{Comparison of the best solutions (\textbf{Best}).}
\label{fig:1}
\end{subfigure}
\hfil

\begin{subfigure}{.7\textwidth}
\begin{tikzpicture}[scale=0.72]
\begin{axis}[
  axis x line=center, axis y line=none, xmin=1, xmax=4, ymin=-3.5, ymax=0, scale only axis, height=4\baselineskip, width=\axisdefaultwidth, ticklabel style={anchor=south, yshift=1.33*\pgfkeysvalueof{/pgfplots/major tick length}, font=\small}, every tick/.style={yshift=.5*\pgfkeysvalueof{/pgfplots/major tick length}}, axis line style={-}, title style={yshift=\baselineskip}, xtick={1,2,3,4}, clip=false
]

\draw[semithick, rounded corners=1pt] (axis cs:1.4166666666666667, 0) |- (axis cs:1.0, -1.5) node[font=\small, fill=white, inner xsep=5pt, outer xsep=-5pt, anchor=east] {HCSA-MDS};

\draw[semithick, rounded corners=1pt] (axis cs:2.4523809523809526, 0) |- (axis cs:1.0, -2.5) node[font=\small, fill=white, inner xsep=5pt, outer xsep=-5pt, anchor=east] {HBA};

\draw[semithick, rounded corners=1pt] (axis cs:2.5714285714285716, 0) |- (axis cs:4.0, -2.5) node[font=\small, fill=white, inner xsep=5pt, outer xsep=-5pt, anchor=west] {SAMDS};

\draw[semithick, rounded corners=1pt] (axis cs:3.5595238095238093, 0) |- (axis cs:4.0, -1.5) node[font=\small, fill=white, inner xsep=5pt, outer xsep=-5pt, anchor=west] {HGA-MDS};
\end{axis}
\end{tikzpicture}

\caption{HCSA-MDS worst solutions (\textbf{Worst}) compared to HBA,SAMDS, and HGA-MDS best solutions.}
\label{fig:2}
\end{subfigure}
\end{adjustbox}

\hfil

\begin{adjustbox} {width=.5\columnwidth,center}
\begin{subfigure}{.7\textwidth}
\begin{tikzpicture}[scale=0.72]
\begin{axis}[
  axis x line=center, axis y line=none, xmin=1, xmax=4, ymin=-3.5, ymax=0, scale only axis, height=4\baselineskip, width=\axisdefaultwidth, ticklabel style={anchor=south, yshift=1.33*\pgfkeysvalueof{/pgfplots/major tick length}, font=\small}, every tick/.style={yshift=.5*\pgfkeysvalueof{/pgfplots/major tick length}}, axis line style={-}, title style={yshift=\baselineskip}, xtick={1,2,3,4},clip=false
]

\draw[semithick, rounded corners=1pt] (axis cs:1.4047619047619047, 0) |- (axis cs:1.0, -1.5) node[font=\small, fill=white, inner xsep=5pt, outer xsep=-5pt, anchor=east] {HCSA-MDS};

\draw[semithick, rounded corners=1pt] (axis cs:1.7738095238095237, 0) |- (axis cs:1.0, -2.5) node[font=\small, fill=white, inner xsep=5pt, outer xsep=-5pt, anchor=east] {HBA};

\draw[semithick, rounded corners=1pt] (axis cs:3.3452380952380953, 0) |- (axis cs:4.0, -2.5) node[font=\small, fill=white, inner xsep=5pt, outer xsep=-5pt, anchor=west] {HGA-MDS};

\draw[semithick, rounded corners=1pt] (axis cs:3.4761904761904763, 0) |- (axis cs:4.0, -1.5) node[font=\small, fill=white, inner xsep=5pt, outer xsep=-5pt, anchor=west] {SAMDS};

\draw[ultra thick, line cap=round] (axis cs:1.4047619047619047, -0.75) -- (axis cs:1.7738095238095237, -0.75);

\end{axis}
\end{tikzpicture}
\caption{Standard deviation (\textbf{std.})}
\label{fig:3}
\end{subfigure}

\end{adjustbox}
\caption{Critical Difference plots evaluating HCSA-MDS, HBA, SAMDS, and HGA-MDS for the first benchmark.}
\label{figure:Cdplots}
\end{figure}

\begin{figure}
\centering
  \includegraphics[width=20cm,height=22cm,keepaspectratio]{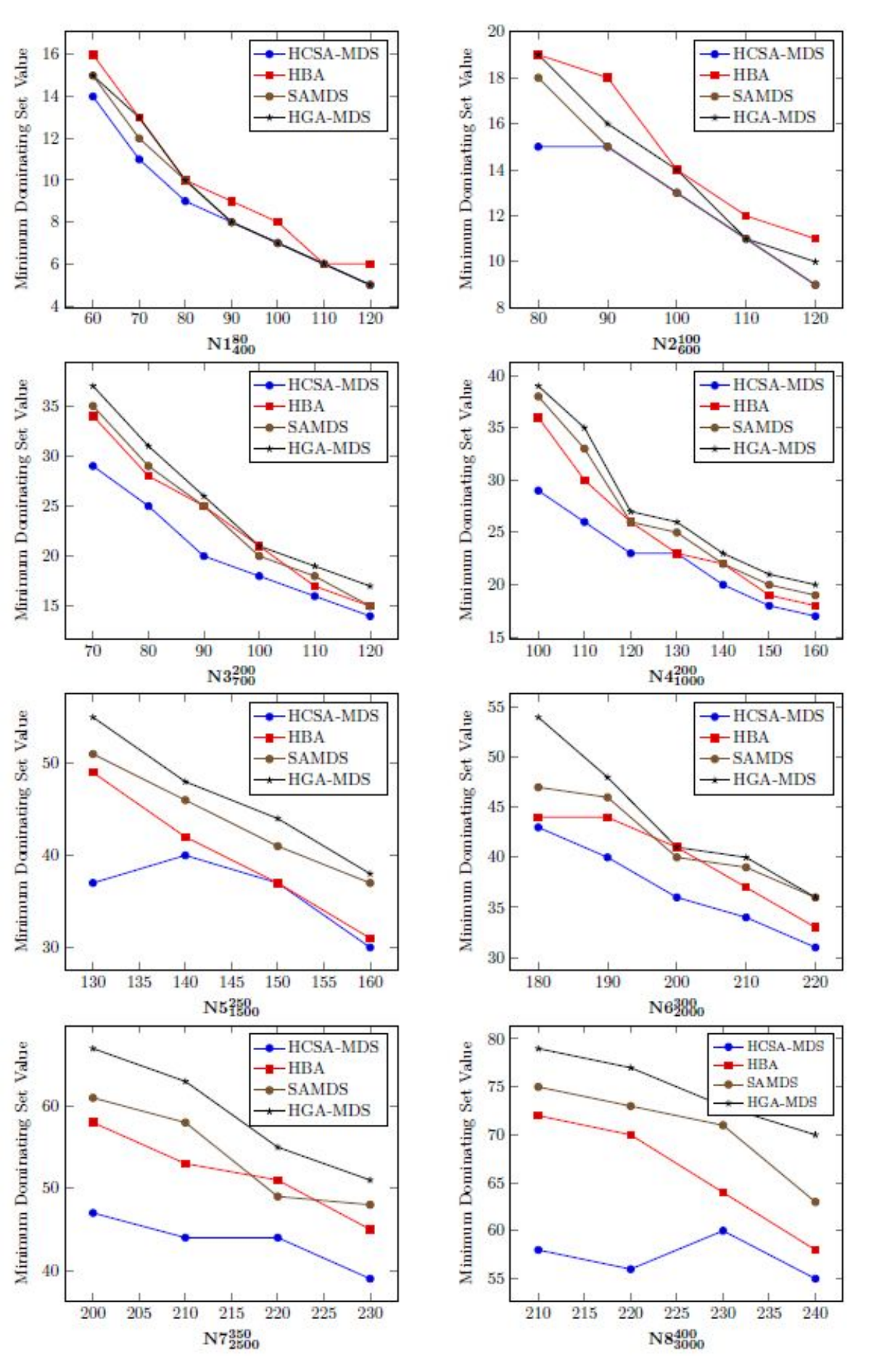}
\caption{Best solution (\textbf{Best}) analysis of various approaches.}
\label{fig:bestgraphs}
\end{figure}

 \textbullet\ HCSA-MDS outperforms all the state-of-the-art solutions in terms of providing the best solution. Furthermore, HCSA-MDS can improve the best-known solution in 32 out of 42 graph instances and match the best-known solution for the rest of the 10 instances. For the larger graphs, the differences between HCSA-MDS and the other algorithms begin to grow. For example, in instances with 300, 350, and 400 vertices, HCSA-MDS provided better solutions in all graph instances with a substantial difference, as shown in Figure \ref{fig:bestgraphs}, which simulated the results of a best-solution comparison between HCSA-MDS and the other approaches. In addition, we used a critical difference diagram in Figure \ref{figure:Cdplots}\subref{fig:1} to verify these results, showing that HCSA-MDS beats all other algorithms, followed by HBA, SAMDS, and HGA-MDS, which is the worst. In summarizing, we can assert that HCSA-MDS beats the state of the art in terms of solution quality by a large margin.\\

 \textbullet\ Concerning the worst solution obtained over 10 runs, in 23 out of 42 graph instances, HCSA-MDS can outperform the currently best-known approaches. Furthermore, throughout the ten runs, HCSA-MDS does not produce worse solutions than those produced by the best solution in the state of the art. HCSA-MDS improves the best solution obtained by HBA in 29 instances and only matches HBA's performance in 11 graphs. In only two cases did HBA find a solution that was better than HCSA-MDS's worst solution. In 32 out of 42 graph instances, the HCSA-MDS worst solution is better than the best solution produced by SAMDS, and in 8 cases, it matches the best solution. In only two cases, the best solution provided by SAMDS is better than the worst solution by HCSA-MDS. In 35 instances, HCSA-MDS gave worst solutions that were better than HGA-MDS's best solutions, 6 solutions were similar to the best solution, and in only one case, HGA outperformed HCSA-MDS in terms of worst solution. To recap, these details can be found in the Ranks Table \ref{table:Ranks_t}. Table \ref{table:worstvsbest} displays Wilcoxon's test statistics which show that there is a significant difference between HCSA-MDS and other approaches. Finally, Figure \ref{figure:Cdplots}\subref{fig:2} shows the critical difference plot where HCSA-MDS exceeds the state of the art in terms of the worst solution when compared to the best solution yielded by literature approaches.
\begin{table}[H]\centering
\bgroup
\def\arraystretch{1.5}
\begin{tabular}{c|ccc|c|}
\cline{2-5}
& \multicolumn{3}{|c|}{HCSA-MDS} & N\textsuperscript{o} \\ \cline{1-4}
\multicolumn{1}{|c|}{$\longrightarrow $} & $>$ & $<$ &
$=$ &  instances \\ \hline
\multicolumn{1}{|c|}{HBA} & 29 & 2 & 11 & 42 \\
\multicolumn{1}{|c|}{SAMDS} & 32 & 2 & 8 & 42 \\
\multicolumn{1}{|c|}{HGA-MDS} & 35 & 1 & 6 & 42 \\ \hline
\end{tabular}
\egroup
\caption{The ranks in summary.}
\label{table:Ranks_t}
\end{table}

\begin{table}[H]\centering
\bgroup
\def\arraystretch{1.5}
\begin{tabular}{|c||c||c|c|}
\hline
Criteria & Algorithm & $Z$ & Sig. ($p$ value) \\ \hline
Worst of HCSA-MDS  & HCSA-MDS vs HBA & -4.671 & 0.000 \\ \cline{2-4}
VS & HCSA-MDS vs SAMDS & -4.918 & 0.000 \\ \cline{2-4}
Best of Others & HCSA-MDS vs HGA-MDS & -5.192 & 0.000 \\ \hline
\end{tabular}
\egroup
\caption{Results of statistical significance between HCSA-MDS's worst solutions and other approaches best solutions.}
\label{table:worstvsbest}
\end{table}

  \textbullet\ The results of the standard deviation (column \textbf{std.}) in Table \ref{table:firstbenchmark} clearly show that HCSA-MDS is more stable than HBA, as it gave stable dominating set values in 10 out of 42 instances, whereas HBA only provided stable dominating set values in 6 out of 42 instances. For both approaches, the standard deviation for the other cases is too near zero. SAMDS, on the other hand, outperforms HGA-MDS, which is the worst, especially in the densest graphs. This result is verified by the mean ranks of critical difference plots in Figure \ref{figure:Cdplots}\subref{fig:3}, where HCSA-MDS beats the other algorithms with statistical significance. Despite being ranked second, HBA's stability is statistically equal to that of HCSA-MDS. Eventually, the HGA-MDS approach is the weakest in the analysis.

\subsubsection{Numerical results for the HCSA-MDS and various approaches on the second benchmark}
We discuss the effectiveness of the suggested HCSA-MDS on the second test case benchmark in this section. Our method's results are evaluated against those of state-of-the-art algorithms. Hence, we compare HCSA-MDS against a variety of greedy algorithms from \cite{sanchis2002experimental}, where Sanshis L described and studied many greedy heuristics designated Greedy, GreedyRev, GreedyRan, GreedyVote, and GreedyVoteGr, which is an updated variant of GreedyVote. According to the evaluations, GreedyVoteGr beats other greedy heuristics when employing the local search process.\\

Similarly, we compare HCSA-MDS with the Hybrid Genetic Algorithm abbreviated HGA-MDS from \cite{hedar2012simulated}, as well as Simulated Annealing (SAMDS) from the same paper \cite{hedar2012simulated} which shows better performance than HGA-MDS \mbox{and the greedy heuristics.}\\

The numerical values are presented in Tables \ref{table:N400} and \ref{table:N800}. For each graph instance, the results of various approaches are presented in terms of the average solution (\textbf{Avg.}) obtained in 10 runs in each table. Three alternative graph densities are taken: from dense to sparse  0.5,0.3 and 0.1. For today's state-of-the-art approaches, such benchmark sets are just very simple. Thus, we examine how many run out of 10, the Minimum Dominating Set was reached, which is represented in the column labeled "\textbf{Opt. Rea.}".\\

The experiments in Tables \ref{table:N400} and \ref{table:N800}, show that HCSA-MDS is the highest performing algorithm with absolute distinctions in terms of average solution and reaching the optimal solution. Moreover,almost all instances 100\% reaching the optimal in all 10 runs except two instances ($\mathbf{N_{0.3,3}^{400}}$,$\mathbf{N_{0.5,3}^{800}}$) optimal reached 8, 9 \mbox{times respectively out of ten.}\\

To determine the relevance of the differences between the approaches results and HCSA-MDS we used Wilcoxon's signed-rank test. it only makes sense to consider only GreedyVoteGr, HGA-MDS, and SAMDS approaches, as they outperform the other greedy algorithms, moreover the results of the rest of the greedy methods are too similar.\\

The Test statistics presented in Table \ref{table:wilcoxon2}, show that there is a significant difference between HGA-MDS and the best performing algorithm in the Greedy series from \cite{sanchis2002experimental}, which is GreedyVoteGr in terms of both criteria. Furthermore, HGA-MDS could perform equally as well as SAMDS in terms of the mean of the solutions found, although SAMDS outperforms HGA-MDS in terms of the attainment rate of reaching the Minimum Dominating Set. Finally, HCSA-MDS outperforms SAMDS, HGA-MDS, and GreedyVoteGr since it is outperformed by HGA-MDS. To conclude, we can state that HCSA-MDS outperforms the state-of-the-art approaches in terms of robustness and stability.

\begin{table}[H]
\begin{adjustbox}{width=1.4\linewidth,center}
\begin{tabular}{ll|llllllllllllllll}
\hline
&  & \multicolumn{2}{|l}{Greedy} & \multicolumn{2}{l}{GreedyRev} &
\multicolumn{2}{l}{GreedyRan} & \multicolumn{2}{l}{GreedyVote} &
\multicolumn{2}{l}{GreedyVoteGr} & \multicolumn{2}{l}{HGA-MDS} &
\multicolumn{2}{l}{SAMDS} & \multicolumn{2}{l}{HCSA-MDS} \\ \cline{3-18}
&  &  & Opt. &  & Opt. &  & Opt. &  & Opt. &  & Opt. &  & Opt. &  & Opt. &
& Opt. \\
Graph & d & Avg. & Rea. & Avg. & Rea. & Avg. & Rea. & Avg. & Rea. & Avg. & Rea. &
Avg. & Rea. & Avg. & Rea. & Avg. & Rea. \\ \hline
N$_{0.1,d}^{400}$ & 8 & 14.2 & 0 & 16.1 & 4 & 30.9 & 0 & 9.2 & 4 & 8 & 10 & 8
& 10 & 8 & 10 & 8 & 10 \\
& 11 & 22.1 & 0 & 25.2 & 0 & 33.2 & 0 & 16.8 & 0 & 15.6 & 5 & 11.1 & 9 & 11.1
& 9 & 11 & 10 \\
& 14 & 24 & 0 & 25.6 & 0 & 35.2 & 0 & 19 & 1 & 18.6 & 3 & 14.4 & 6 & 14.2 & 9
& 14 & 10 \\
& 18 & 27.3 & 0 & 27.3 & 0 & 38.5 & 0 & 21.8 & 0 & 20.4 & 4 & 18.4 & 6 & 18
& 10 & 18 & 10 \\
& 23 & 31.3 & 0 & 28.8 & 0 & 41.8 & 0 & 24.9 & 0 & 24.8 & 0 & 24.2 & 4 & 23
& 10 & 23 & 10 \\ \hline
N$_{0.3,d}^{400}$ & 3 & 6.8 & 0 & 9.9 & 0 & 10.9 & 0 & 6.8 & 4 & 6 & 4 & 3 &
10 & 3.8 & 8 & 3.2 & 8 \\
& 5 & 9 & 0 & 10.6 & 0 & 11.7 & 0 & 8.7 & 0 & 8.7 & 0 & 5.3 & 7 & 5.2 & 8 & 5
& 10 \\
& 8 & 10.9 & 0 & 11.6 & 0 & 13.3 & 0 & 9.3 & 0 & 9.2 & 0 & 8.1 & 9 & 9 & 8 &
8 & 10 \\
& 11 & 14 & 0 & 12.6 & 0 & 15.6 & 0 & 11.1 & 9 & 11 & 10 & 11 & 10 & 11 & 10
& 11 & 10 \\
& 14 & 16.1 & 0 & 14.2 & 0 & 18.2 & 0 & 14 & 10 & 14 & 10 & 14 & 10 & 14 & 10
& 14 & 10 \\ \hline
N$_{0.5,d}^{400}$ & 3 & 5 & 0 & 6.3 & 0 & 6.4 & 0 & 5 & 0 & 5 & 0 & 3.1 & 9
& 3.1 & 9 & 3 & 10 \\
& 5 & 6 & 2 & 6.6 & 0 & 7 & 0 & 5.6 & 5 & 5.4 & 6 & 5.2 & 8 & 5.3 & 9 & 5 &
10 \\
& 8 & 8.8 & 3 & 8.2 & 8 & 9 & 0 & 8.1 & 9 & 8 & 10 & 8 & 10 & 8 & 10 & 8 & 10
\\
& 11 & 11.9 & 3 & 11 & 10 & 11.9 & 4 & 11 & 10 & 11 & 10 & 11 & 10 & 11 & 10
& 11 & 10 \\ \hline
\end{tabular}
\end{adjustbox}
\caption{Numerical results for various approaches as well as HCSA-MDS on $\mathbf{N_{p,d}^{400}}$.}
\label{table:N400}
\end{table}

\begin{table}[H]
\begin{adjustbox}{width=1.4\linewidth,center}
\begin{tabular}{ll|llllllllllllllll}
\hline
&  & \multicolumn{2}{|l}{Greedy} & \multicolumn{2}{l}{GreedyRev} &
\multicolumn{2}{l}{GreedyRan} & \multicolumn{2}{l}{GreedyVote} &
\multicolumn{2}{l}{GreedyVoteGr} & \multicolumn{2}{l}{HGA-MDS} &
\multicolumn{2}{l}{SAMDS} & \multicolumn{2}{l}{HCSA-MDS} \\ \cline{3-18}
&  &  & Opt. &  & Opt. &  & Opt. &  & Opt. &  & Opt. &  & Opt. &  & Opt. &
& Opt. \\
Graph & d & Avg. & Rea. & Avg. & Rea. & Avg. & Rea. & Avg. & Rea. &
Avg. & Rea. & Avg. & Rea. & Avg. & Rea. & Avg. & Rea. \\ \hline
N$_{0.1,d}^{800}$ & 11 & 26.3 & 0 & 30.5 & 0 & 40.3 & 0 & 23.7 & 0 & 22.8 & 1
& 11.3 & 7 & 11.3 & 7 & 11 & 10 \\
& 14 & 28.1 & 0 & 31.4 & 0 & 41.9 & 0 & 23.5 & 0 & 22.4 & 2 & 14 & 10 & 14 &
10 & 14 & 10 \\
& 22 & 34.9 & 0 & 33.1 & 0 & 48.3 & 0 & 27.3 & 0 & 27.3 & 0 & 22.5 & 5 & 22
& 10 & 22 & 10 \\ \hline
N$_{0.3,d}^{800}$ & 3 & 8.7 & 0 & 12.1 & 0 & 12.5 & 0 & 9 & 0 & 8.8 & 1 & 7.9
& 6 & 7.3 & 6 & 3 & 10 \\
& 5 & 9.7 & 0 & 12.2 & 0 & 13.3 & 0 & 9.4 & 0 & 9.4 & 0 & 6.8 & 5 & 5 & 10 &
5 & 10 \\ \hline
N$_{0.5,d}^{800}$ & 3 & 6 & 0 & 7.3 & 0 & 7 & 0 & 6 & 0 & 6 & 0 & 4.4 & 5 &
3.4 & 8 & 3.2 & 9 \\
& 6 & 7.4 & 2 & 7.8 & 0 & 8.3 & 0 & 6.3 & 7 & 6.2 & 8 & 6.5 & 8 & 6 & 10 & 6
& 10 \\ \hline
\end{tabular}
\end{adjustbox}
\caption{Numerical results for various approaches as well as HCSA-MDS on $\mathbf{N_{p,d}^{800}}$.}
\label{table:N800}
\end{table}

\begin{table}[H]\centering
\bgroup
\def\arraystretch{1.5}
\begin{tabular}{|c|c|c|c|}
\hline
Criteria & Algorithm & $Z$ & Sig. ($p$ value) \\ \hline
Optimal Reached & HGA-MDS \textbf{\textit{vs.}} GrVoteGr & -3.417 & 0.001 \\ \cline{2-4}
(Opt. Rea.) & SAMDS \textbf{\textit{vs.}} HGA-MDS & -2.364 & 0.018 \\ \cline{2-4}
& HCSA-MDS \textbf{\textit{vs.}} HGA-MDS & -3.136 & 0.002 \\ \cline{2-4}
& HCSA-MDS \textbf{\textit{vs.}} SAMDS & -2.716 & 0.007 \\ \hline
Averge & HGA-MDS \textbf{\textit{vs.}} GrVoteGr & -3.413 & 0.001 \\ \cline{2-4}
(Avg.) & SAMDS \textbf{\textit{vs.}} HGA-MDS & -1.609 & 1.08 \\ \cline{2-4}
& HCSA-MDS \textbf{\textit{vs.}} HGA-MDS & -3.158 & 0.002 \\ \cline{2-4}
& HCSA-MDS \textbf{\textit{vs.}} SAMDS & -2.814 & 0.005 \\ \hline
\end{tabular}
\egroup
\caption{The Wilcoxon's signed ranks test results for GreedyVoteGr, HGA-MDS, SAMDS, and HCSA-MDS.}
\label{table:wilcoxon2}
\end{table}

\section{Conclusion}\label{sec:Conclusion}

In this paper, we addressed one of the fundamental NP-Hard problems in graph theory, which is the Minimum Dominating Set problem. We have proposed an effective Hybrid Cuckoo Search Algorithm abbreviated \textbf{HCSA-MDS}. HCSA-MDS outperforms the existing state-of-the-art techniques in terms of the best result obtained,  according to an experimental evaluation on multiple benchmark sets. Furthermore, we tested the robustness of our proposed algorithm to state-of-the-art techniques utilizing the best, average, and even worst solutions, and found that HCSA-MDS outperformed other approaches.\\

HCSA-MDS, in our perspective, outperforms the state-of-art techniques, because it combines the cuckoo search algorithm, which uses L\'{e}vy flights to efficiently explore the search space, with multiple schemes for intensification and genetic crossover operator to exploit new solutions.

\end{document}